\documentclass[12pt,a4paper,final]{iopart}
\pdfoutput=1 

\usepackage{iopams}  
\usepackage{graphicx}
\usepackage[breaklinks=true,colorlinks=true,linkcolor=blue,urlcolor=blue,citecolor=blue]{hyperref}

\expandafter\let\csname equation*\endcsname\relax
\expandafter\let\csname endequation*\endcsname\relax
\usepackage{amsmath}

\allowdisplaybreaks

\begin{document}

\title[]{Thresholds of descending algorithms in inference problems}

\author{Stefano Sarao Mannelli and Lenka Zdeborov\'a}
\address{Institut de Physique Théorique}
\address{CEA, Orme des Merisiers, 91191 Gif-sur-Yvette, France}
\ead{stefano.sarao@ipht.fr, lenka.zdeborova@ipht.fr}

\begin{abstract}
We review recent works \cite{sarao2018marvels,sarao2019passed,sarao2019afraid} on analyzing the dynamics of gradient-based algorithms in a prototypical statistical inference problem. Using methods and insights from the physics of glassy systems, these works showed how to understand quantitatively and qualitatively the performance of gradient-based algorithms. Here we review the key results and their interpretation in non-technical terms accessible to a wide audience of physicists in the context of related works.  

\end{abstract}

\vspace{2pc}

\section{Introduction}

Machine learning as achieved astonishing success across real world problems, such as image classification, speech recognition, text processing, and physical problems, from quantum physics \cite{carleo2017solving}, to astrophysics \cite{ball2010data}, to high-energy physics \cite{radovic2018machine}. Despite these practical successes, a large number of aspects still lacks theoretical understanding. Practitioners identified several prescriptions to construct a working machine learning applications, but it is often unclear why those recipes are effective.
Consider a typical classification task, where a dataset consisting of pictures of cats and dogs is provided to the machine with the correct labels. What follows is the minimization of a cost function. Given new images of pets, the goal of the machine is to be able to correctly classify them into cats and dogs, thus successfully generalizing from what it has seen.

The optimization process itself is puzzling. In general, the cost function is high-dimensional and non-convex. Intuition would suggest that a random initialization would lead to some local spurious, non-informative, minimum with very little hope to achieve a good generalization. Instead, in practice even the use of vanilla gradient descent often leads to good generalization. 
Part of the computer science community analysed the problem geometrically by studying the properties of the cost function \cite{bandeira2016low,kawaguchi2016deep,ge2016matrix,ge2017no,du2017gradient}. They consider generative models where a signal is observed through a noisy channel and it is possible to tune its strength with respect to the strength of the noise, the signal to noise ratio (SNR), and change the landscape. They showed that in a variety of problems, all the minima become equally good or the spurious minima disappear as the SNR becomes sufficiently large, thus making the landscape trivial.

In this work we review the recent effort towards an understanding of the learning dynamics using the tools of disordered systems \cite{sarao2019passed,sarao2019afraid}, and we discuss the difference in performance between message passing algorithms and algorithms for sampling a high-dimensional potential \cite{sarao2018marvels}. The relation between the two approaches becomes apparent from the point of view of Bayesian statistics. 
Let $\pmb X$ be the guess on the hidden signal and $\pmb Y$ the observation, we can express plausible is to observe $\pmb Y$ given our guess, i.e. the likelihood $\mathbb{P}[\pmb Y|\pmb X]$. Bayes formula allows to invert the likelihood into the posterior probability $\mathbb{P}[\pmb X | \pmb Y] \propto \mathbb{P}[\pmb Y | \pmb X]\mathbb{P}[\pmb X]$, that also includes prior information on the guess, such as sparsity or norm constrains. We can write an approximate expression
\begin{equation}\label{eq:posterior}
	\mathbb{P}[\pmb X | \pmb Y] \approx_{\beta\approx1} \frac1{\mathbb{P}[\pmb Y]}\overbrace{\mathbb{P}[\pmb Y | \pmb X]^\beta}^\textit{-cost}\mathbb{P}[\pmb X] \doteq \frac1{\mathcal{Z}[\pmb Y]}e^{-\beta\mathcal{H}(\pmb X, \pmb Y)}\,.
\end{equation}
In the last equality we identify the terms with a Gibbs distribution with inverse temperature $\beta$. Given the posterior we can estimate the signal by considering the expected value:
\begin{equation}
	\hat{\pmb x}_\beta = \mathbb{E}_{\pmb X | \pmb Y;\beta} [\pmb X]\;.
\end{equation}
Observe that when the inverse temperature parameter equals 1, Eq.~(\ref{eq:posterior}) is the posterior probability of the problem. As $\beta$ tends to infinity, the cost dominates and optimizing will maximize the likelihood. 

In the eyes of a statistical physicist, the expected value $\hat{\pmb x}_\beta$ would rather be called $\pmb m$ as it is formally identical to the magnetization of a system under the action of the Hamiltonian. However, the exact computation a this expected value exactly is prohibitive in large dimension, in fact it is as complex as evaluating the partition function. In order to avoid such complication numerous ingenious techniques have been considered in the past to obtain an approximate estimation. Two main approaches consist of approximating the posterior, and sampling the posterior. 

\begin{itemize}
\item The idea of adapting the approximations proposed in disordered systems to computer science problems is not recent, and early works appeared in the 80s and 90s \cite{MPV87,engel2001statistical}. Ideas from physics were transferred to problems in signal processing and optimisation, providing both theoretical understanding and practical algorithms based on Cavity Method and its variations \cite{MM09,REVIEWFLOANDLENKA,LKZ17}. Those methods have the advantage of being at the same time algorithms and analytical tools. In many problems they were proved to be asymptotically optimal \cite{MiolaneUV,LMLKZ17,barbier2017phase}, in the sense that information-theoretically they achieve the best performance in polynomial time. 

\item The best known algorithms that sample the posterior are Monte Carlo and the Langevin algorithm. Studies on the Langevin algorithm in disordered systems have their root in the late 70s \cite{martin1973statistical,De78,kirkpatrick1987dynamics,CHS93}.  Despite the dynamics was understood for some recurrent neural networks in long-time regime \cite{sompolinsky1988chaos,coolen2000statistical}, generalizing and solving the corresponding equations is very difficult even in the simplest models of statistical inference \cite{agoritsas2018out}. Consequently, analysis of the performance of gradient-based algorithms such as the Langevin algorithm remains an open problem. A progress on this question was recently made in a series of works \cite{sarao2018marvels,sarao2019passed,sarao2019afraid} that we review here.
\end{itemize}

The paper is organized as follows: in Section~\ref{sec:model} we introduce the model, in Section~\ref{sec:langevin} we propose a comparison between sampling algorithms and approximate algorithms, in Section~\ref{sec:GF} the gradient flow algorithm is analyzed and compared with the energy landscape.

\section{Spiked matrix-tensor model}\label{sec:model}


\begin{figure}[htb!]
 \centering
 \includegraphics[scale=.45]{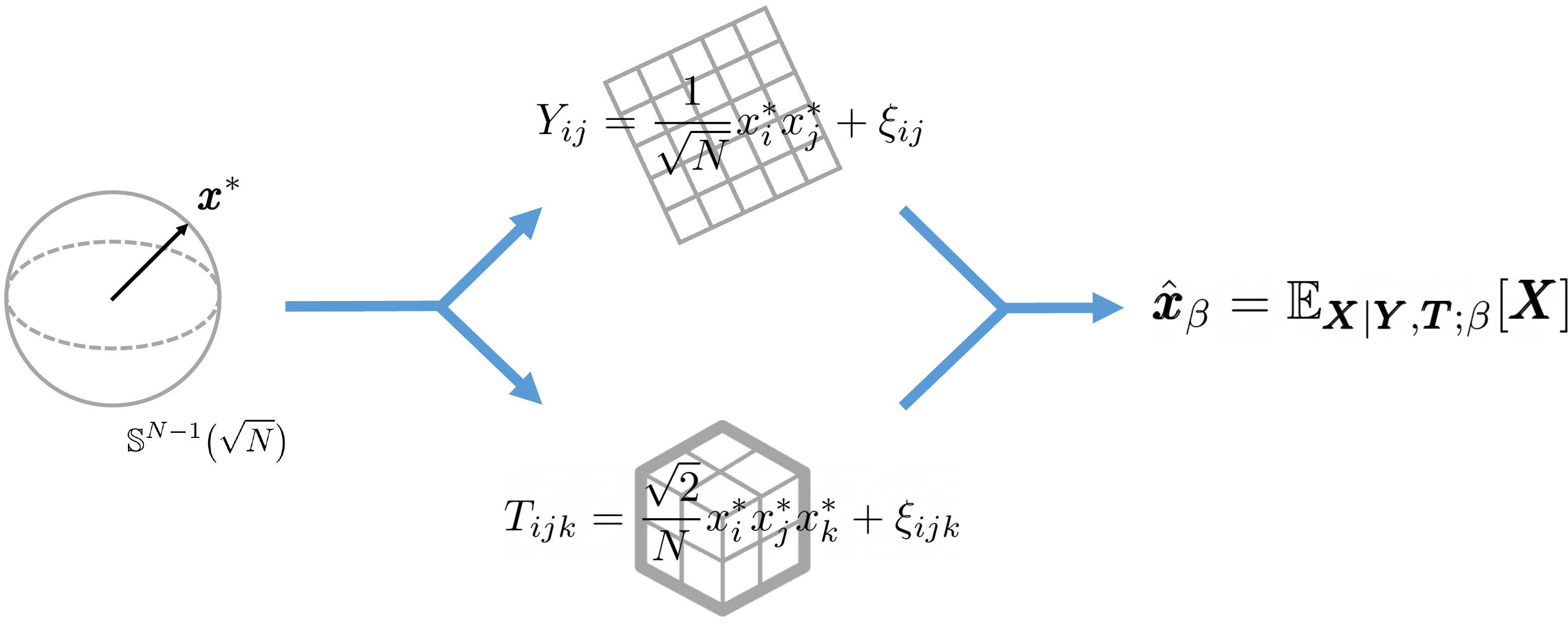}
 \caption{\label{fig:matrix-tensor}Cartoonish representation of the generative process. The left part of the image represents the teacher who samples the ground truth $\pmb x^*$ and uses it to generate the two observations that contain noise, $\xi_{ij}\sim\mathcal{N}(0,\Delta_2)$ and $\xi_{ijk}\sim\mathcal{N}(0,\Delta_3)$. The student, on the right part of the figure, receives the observations and constructs an estimation of the signal, $\hat{\pmb x}_\beta$, by computing the expected value.}
\end{figure}

The model that we study in this report is the \textit{spiked matrix-tensor model}, known in physics as a planted version of the spherical mixed $p$-spin model \cite{gross1984simplest,CHS93,CL04}. Planting is a technique  introduced to study statistical inference and learning problems using the same methods as for their optimization counterparts \cite{krzakala2009hiding}. Planting appears as an additional ferromagnetic bias towards a planted solution (or ground-truth) in the Hamiltonian. In its application to inference, planting permits to introduce a signal, the ground-truth solution, in the formulation of the problem. In the neural-network-learning language this formulation is called \textit{teacher-student scenario}: the \textit{teacher} knows the ground-truth and uses it to generate data, the \textit{student} has to use the data to infer the ground-truth.

The \textit{spiked matrix-tensor model} was introduced in \cite{sarao2018marvels,sarao2019passed,sarao2019afraid} in order to build an inference problem for which the behaviour of the gradient-based dynamics is exactly solvable. 
For the sake of simplicity we will consider $p=3$, which means that the \textit{teacher} samples the ground truth $\pmb x^*$ and generates the data, a matrix and a order 3 tensor. The process is noisy and the data that the student receives, $Y_{ij}$ and $T_{ijk}$, have an intrinsic Gaussian noise of variance $\Delta_2$ and $\Delta_3$ respectively. The two observations are rescaled in order to have an extensive free energy in the size of the system $N$. The generative process is represented in Fig.~\ref{fig:matrix-tensor}. Substituting the data into the posterior Eq.~(\ref{eq:posterior}) and absorbing constant terms into the pre-factor, we obtain the Hamiltonian
\begin{equation}\label{eq:Hamiltonian}
	\begin{split}
		\mathcal{H} &= -\frac1{\Delta_2\sqrt{N}}\sum_{i<j}\xi_{ij}x_ix_j -\frac{\sqrt{2}}{\Delta_3N}\sum_{i<j<k}\xi_{ijk}x_i x_j x_k - \frac{N}{2\Delta_2}m^2 - \frac{N}{3\Delta_3}m^3\,,
	\end{split}
\end{equation}
where $m = \frac1N\sum_i x_i x_i^*$ is the overlap with the signal. Observe that the noise terms $\xi_{ij}$ ($\xi_{ijk}$) in the equation are rescaled by $\sqrt{N}$ ($N$ respectively) in order to have a problem that is neither impossibly hard (very high noise) nor trivially easy (very small noise). Under this choice of scaling of noise, we observe different transitions for values of $\Delta_2$ and $\Delta_3$ of order $O(1)$.

The spiked matrix-tensor model is a natural candidate for our analysis as it has high-dimensional non-convex energy landscape. 
The \textit{algorithmic transition}, after which algorithms start to detect the signal, occurs at the same noise scaling as the \textit{information-theoretic transition} for detection.
The model is analytically tractable using different methods allowing to experiment and compare.
We remark that the spiked tensor model does not have an algorithmic and information-theoretic transitions occurring in the same scaling regime of the noise, thus it is a less interesting candidate for our analysis. 

\section{Sampling algorithms vs approximate algorithms}\label{sec:langevin}

\begin{figure}[htb!]
 \centering
 \includegraphics[scale=.6]{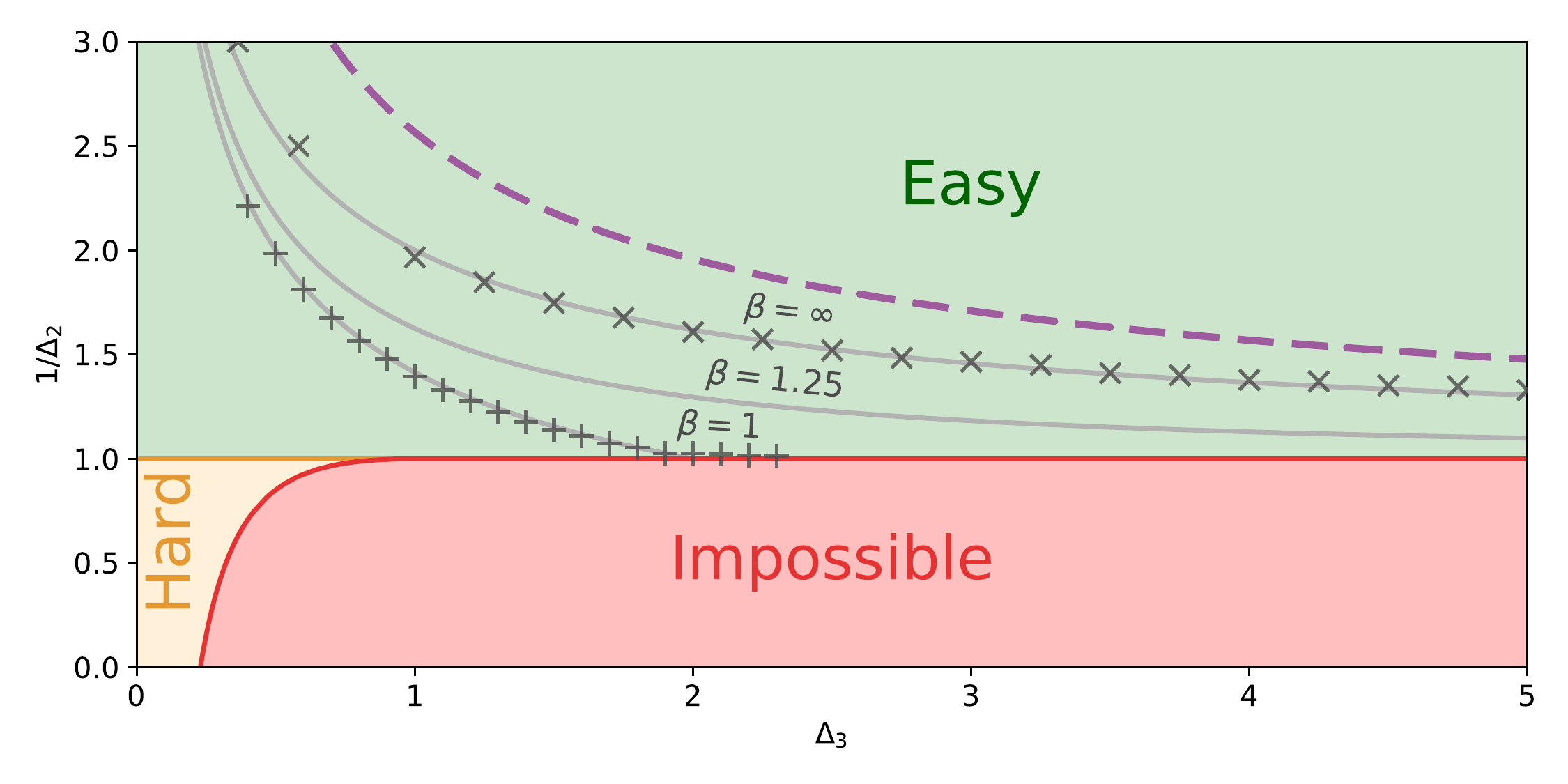}
 \caption{\label{fig:phase_diagram_T1}Phase diagram of the spiked matrix-tensor model with $p=3$. As the variances of the noise in matrix, $\Delta_2$, and the noise in the tensor, $\Delta_3$, change different phases appear. We can distinguish the \textit{easy} (green) phase where AMP can detect the signal, the \textit{impossible} (red) phase where it is information-theoretically impossible to detect the signal, and the \textit{hard} (yellow) phase the where the signal can in principle be detected, but it is expected to take exponential time as it requires to jump over an energy barrier. The grey lines in the easy phase represent the algorithmic transition of the Langevin algorithm for $\beta = 1$, $\beta = 1.25$, and $\beta = \infty$. For a fixed $\beta$, the Langevin algorithm starts to detect the signal above the respective grey line. The plus marks and the cross marks are the extrapolation of the Langevin threshold from the numerical study of the dynamical equations. We can observe good agreement with the analytical prediction. The purple dashed line characterizes the trivialization transition, above that line the energy landscape does not present any spurious minima.}
\end{figure}

We are going to consider an algorithmic version of cavity method as an example of an approximation algorithm.\cite{MM09}. This algorithm was developed independently in the information theory and Bayesian inference community under the name of belief propagation \cite{gallager1962low,pearl1982reverend,pearl1986fusion}. In the case of fully connected models, belief propagation can be simplified by assuming a Gaussian structure in the beliefs, leading to the Approximate Message Passing (AMP) algorithm \cite{DMM09,LKZ17}. AMP presents numerous remarkable features: it provably achieves optimal performances in many problems including the spiked matrix-tensor \cite{MiolaneUV,LMLKZ17,barbier2017phase,sarao2018marvels} and its average behaviour can be analytically followed by a set of equations called \textit{state evolution} \cite{javanmard2013state}. State evolution equations allow to portrait the phase diagram of this model, see Fig.~\ref{fig:phase_diagram_T1}, it was done in \cite{sarao2018marvels} generalizing the results of \cite{LMLKZ17} on the spiked tensor model. The phase diagram can now be used as a baseline for the behaviour of the sampling algorithms.

\bigskip

In order to sample from the posterior probability it is necessary to design a dynamics that has the posterior probability as its stationary measure at large times. A typical sampling algorithm with this objective is the Langevin algorithm.
Given a Hamiltonian $\mathcal{H}$ of a spherical system, Langevin dynamics describes the evolution of the system coupled with a thermal bath at temperature $T = 1/\beta$
\begin{equation}
	\frac{d}{dt} x_i(t) = -\frac{\partial\mathcal{H}(\pmb x)}{\partial x_i}(t) - \mu(t) x_i(t) + \eta_i(t)
\end{equation}
where $\mu(t)$ is a Langrange multiplier that imposes the spherical constraint and $\eta_i(t)$ is the Langevin noise with $\mathbb{E}[\eta_i(t)] = 0$ and $\mathbb{E}[\eta_i(t)\eta_j(t')] = \frac2\beta \delta_{ij}\delta(t-t')$.
In the late 70s, techniques \cite{De78} for the study of Langevin dynamics were adapted to disordered systems providing a set of PDEs on the evolution of few relevant observables. More recently, the results of these techniques have been proved with mathematical rigour in the mixed $p$-spin model \cite{arous2006cugliandolo,dembo2019dynamics}. Those methods have been generalized to the study of planted systems \cite{cammarota2012aging} and applied to the present problem in \cite{sarao2018marvels, sarao2019afraid}. Two variants of the dynamical mean field theory were used to derive the corresponding equations: the dynamical cavity method was used in \cite{sarao2018marvels},  and the generating functional formalism \cite{sarao2019afraid}. The equations obtained characterize the evolution of: the alignment of the system with the ground truth $m(t) = \lim_{N\rightarrow\infty} \frac1N \mathbb{E}_{\xi,\eta}\sum_i x_i(t) x_i^*$, the self-alignment at different times $C(t,t') = \lim_{N\rightarrow\infty} \frac1N \mathbb{E}_{\xi,\eta}\sum_i x_i(t) x_i(t')$, and the response to a perturbation of the Hamiltonian at a previous time $R(t,t') = \lim_{N\rightarrow\infty} \frac1N \mathbb{E}_{\xi,\eta}\sum_i \delta x_i(t)/\delta \eta_i(t')$. 
\begin{align}
	\begin{split}
		&\frac{\partial}{\partial t} C(t,t') = - \mu(t)\,C(t,t')+
		Q'(m(t)) m(t') + \int_0^t
		R(t,t'')Q''(C(t,t''))C(t',t'') dt''
		\\
		&\quad+ \int_0^{t'} R(t',t'')Q'(C(t,t'')) dt''\,,
	\end{split}\label{eq:CHSCK_C}
	\\
	\begin{split}
		&\frac{\partial}{\partial t} R(t,t')= - \mu(t)\,R(t,t') +\int_{t'}^t R(t,t'')Q''(C(t,t''))R(t'',t') dt''\,,
	\end{split}\label{eq:CHSCK_R}
	\\
	\begin{split}
		&\frac{d}{d t} m(t) = -\mu(t)\,m(t)+Q'(m(t)) + \int_{0}^t R(t,t'')m(t'') Q''(C(t,t'')) dt''\,,
	\end{split}\label{eq:CHSCK_Cbar}
\end{align}
with $Q(x) = \frac{x^2}{2\Delta_2} + \frac{x^3}{3\Delta_3}$, the initial conditions $R(t,t')=0$  for all $t<t'$,
$\lim_{t'\rightarrow t^-}R(t,t')=1\;\forall t$ and $C(t,t)=1\;\forall t$ that allows to derive and additional equation for $\mu(t)$. The spiked matrix-tensor model has the nice feature of having a closed form for these equations, allowing an easier evaluation of the numerical solution by propagation from the initial conditions. In \cite{sarao2018marvels,sarao2019passed} the limits of Langevin and gradient descent (respectively) have been evaluated numerically by extrapolation from the numerical solutions, see Fig.~\ref{fig:phase_diagram_T1}. In general the dynamical equations do not close, thus a self-consistent loop is necessary in order to evaluate a numerical solution limiting the times accessible in the numerics \cite{roy2019numerical}. 

An alternative can be derived from the work \cite{cugliandolo1993analytical} where the authors proposed an ansatz for the large time behaviour of the $p$-spin model, which assumes two time scales. The authors also showed that the dynamics is attracted by states - called \textit{threshold states} - characterized by a Hessian that displays marginality, i.e. its spectrum touches the zero. In \cite{sarao2019afraid}, these two ideas are used to derive the analytical threshold of the Langevin dynamics and gradient descent, by assuming that initially the dynamics will tend to the threshold states and at later times it will increase the alignment with the ground truth. The growth is exponential and the exponent is $\Lambda(\Delta_2,\Delta_3;\beta) = \frac1{\Delta_2}-\sqrt{\frac1{\Delta_2}+\frac{2(1-\Delta_2/\beta)}{\Delta_3}}$, the phase transition occurs when the exponent crosses the null value. Analytical and numerical results are shown in Fig.~\ref{fig:phase_diagram_T1} giving a perfect agreement. 

\begin{figure}[htb!]
 \centering
 \includegraphics[scale=.65]{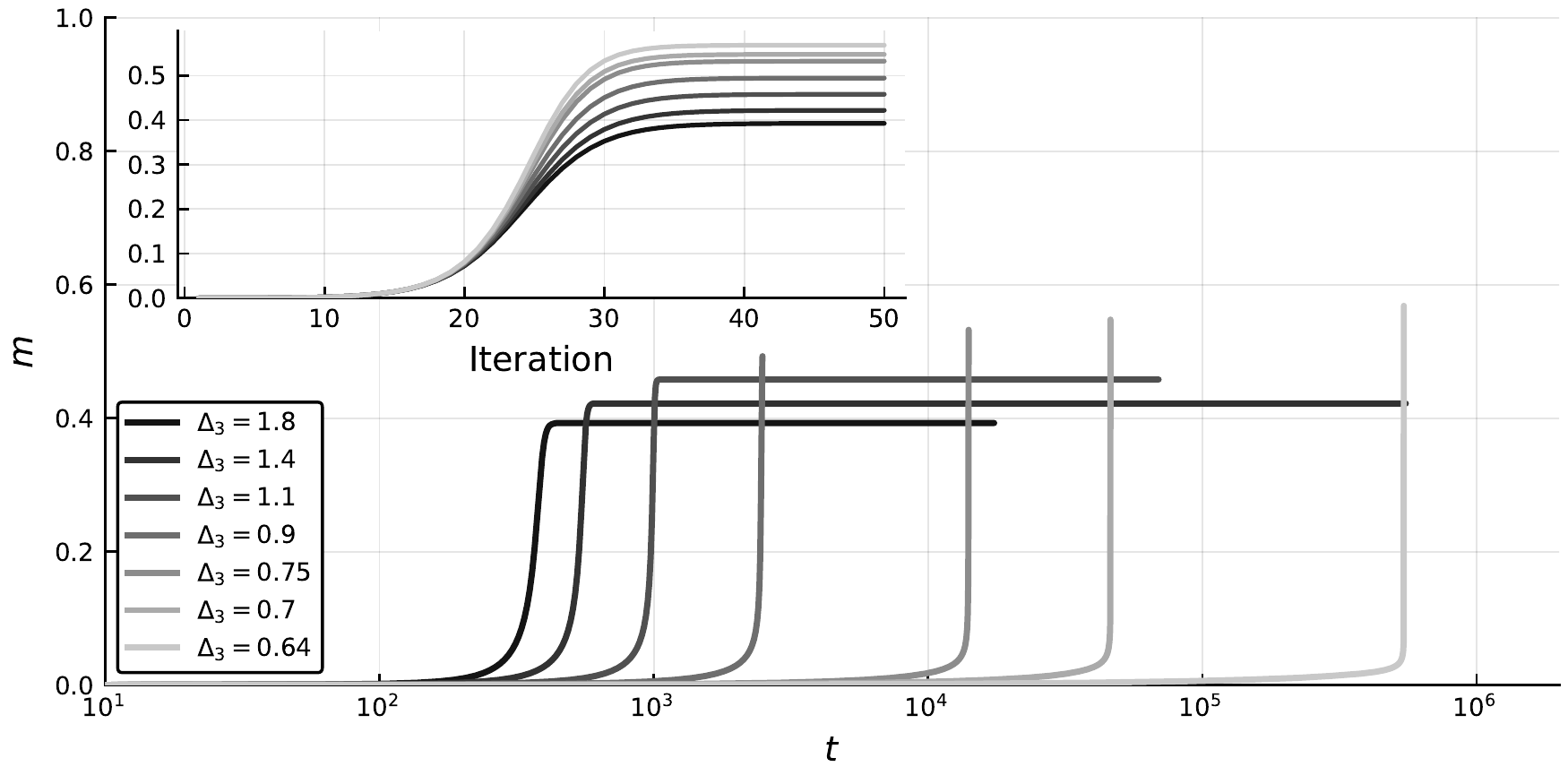}
 \caption{\label{fig:evolution_comparison}Comparison of the evolution of the overlap with the signal in Langevin dynamics and AMP (inset), for $\Delta_2 = 0.70$ and several values of $\Delta_3$.}
\end{figure}

The results suggest that sampling algorithms have worse algorithmic threshold than AMP. This idea was foreseen in \cite{antenucci2018glassy}, where the authors used a large deviation analysis \cite{monasson1995structural} to find exponentially many atypical glassy states in the landscape. They conjectured that the presence of this atypical glassy states may block the dynamics of sampling algorithms. The same analysis was also performed in the spiked matrix-tensor model confirming their findings \cite{sarao2018marvels}.

Another signature of the different transitions appears in the evolution of AMP and Langevin dynamics, Fig.~\ref{fig:evolution_comparison}. For a fixed value of $\Delta_2$ (with $\Delta_2<1$), we can compare evolutions for different values of $\Delta_3$. As the system gets closer to the transition, the time to find the transition increases. We can thus observe that AMP maintains the same typical time to find the solution for the different values of $\Delta_3$, instead the typical time of the Langevin dynamics increases exponentially as $\Delta_3$ becomes smaller. This illustrates the counter-intuitive finding that making the problem simpler by decreasing the noise in the tensor actually harms the Langevin evolution.

\section{Gradient flow and geometry}\label{sec:GF}

\begin{figure}[htb!]
 \centering
 \includegraphics[scale=.55]{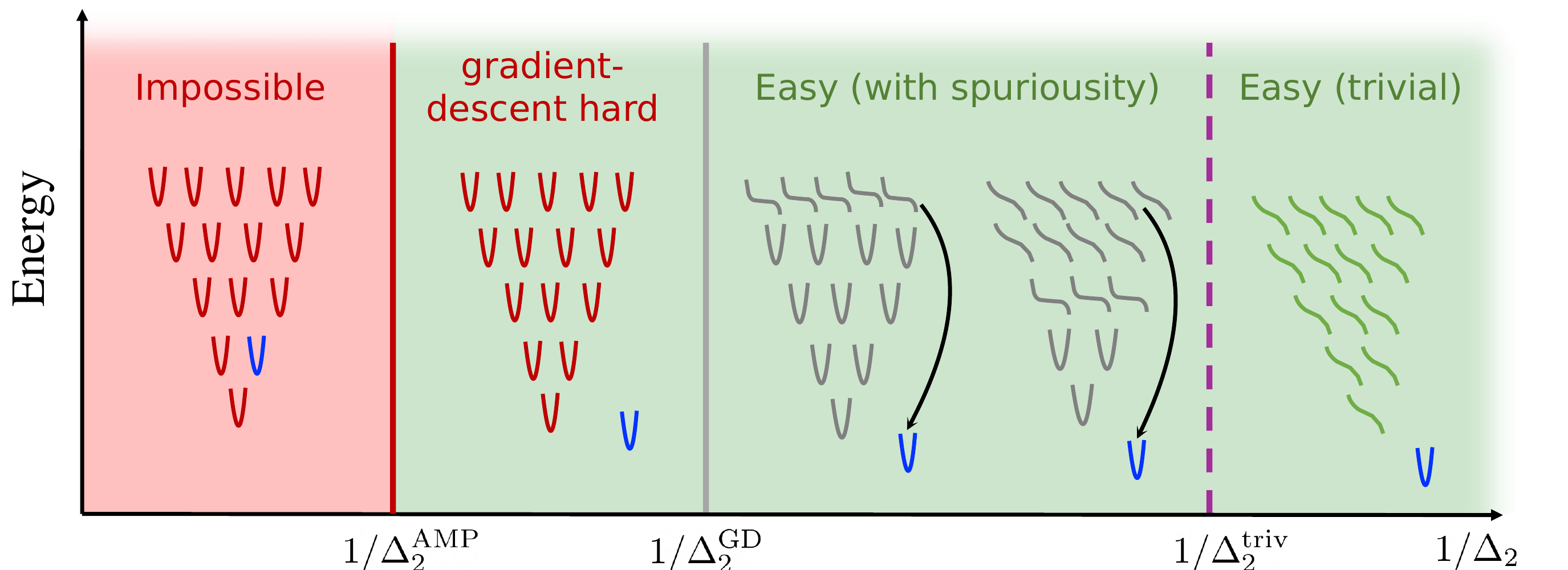}
 \caption{\label{fig:dynamics_sketch}The cartoon represents the energy landscape for an arbitrary value of $\Delta_3>1$. The good minimum is drawn in blue. $1/\Delta_2$ plays the role of the SNR. Starting from low SNR, in the impossible region it is thermodynamically impossible to distinguish between good and bad minima. Increasing the SNR, the good minimum becomes energetically favored but the exponential number of the spurious minima stops the dynamics. At larger SNR the threshold minima becomes saddles pointing toward the good minimum and gradient descent starts to find the solution. Finally the SNR becomes larger than the trivialization threshold and only the good minimum survives.}
\end{figure}
It was already clear in \cite{cugliandolo1993analytical} that $\beta$  enters in a smooth way in the dynamical equations, thus studying the limit $\beta\rightarrow\infty$ we can derive the behaviour of gradient descent dynamics. In machine learning
gradient descent and its several variations (e.g. stochastic gradient descent) are usually used to minimize the cost function. Currently very few problems are amenable to analytical analysis of the dynamics.

In the 80s \cite{bray1981metastable} and in the early 2000 \cite{crisanti2003complexity,cavagna2003formal,crisanti2004quenched} there was an effort to understand the geometrical structure of the energy landscape in disordered models. Given the number of critical points of the model, $\mathcal N_c$, the studies focused on the annealed (and quenched) complexity defined as $\log \mathbb E[\mathcal N_c]$ (and $\log \mathbb E[\mathcal N_c]$, respectively). The authors used an expression that enumerates the number of critical points, namely the \textit{Kac-Rice formula} \cite{adler2009random}, computed using replica theory. Recently another approach for
the evaluation of the Kac-Rice formula has been proposed that uses random matrix theory, giving fruitful results in the $p$-spin model (planted and unplanted) \cite{auffinger2013random,arous2017landscape,ros2018complex}.
In \cite{sarao2019passed} the analysis was generalized to the spiked matrix-tensor model allowing to distinguish between regions where exponentially many minima are present and regions where only the good minima appear. The line that separates them is the \textit{trivialization transition} line. As gradient descent is run above this line, provided that the time-discretization is thin enough, we have a guarantee of finding the good minimum. Many papers \cite{kawaguchi2016deep,ge2016matrix,ge2017no,du2017gradient} have focused on this aspect showing in several problems that as the SNR is strong enough the bad minima disappear, or all the minima become equally good. In the spiked matrix-tensor model geometrical trivialization and gradient descent transition can both be pinpointed on the phase diagram. The results, Fig.~\ref{fig:phase_diagram_T1}, show that gradient descent starts to detect the signal before the trivialization transition has occurred. Although the algorithmic threshold of gradient descent can not occur after the trivialization transition, it might appear counter-intuitive to understand why the two lines do not coincide and there is a distance of order $\mathcal{O}(1)$ that separates them. In \cite{sarao2019afraid} the puzzle was solved, Fig.~\ref{fig:dynamics_sketch}. The authors showed that, moving from a low SNR region where the algorithm fails, the algorithmic transition of gradient descent appears when the dominant minima (the threshold states or threshold minima) develop an instability, a Baik-BenArous-P\'ech\'e instability \cite{baik2005phase}, becoming saddles with a single negative direction that points toward the signal. In this region there are still exponentially many minima that do not carry information on the signal, nevertheless the dynamics is first attracted by the saddles at the threshold that shield the system from the bad minima and point in the right direction.

\section{Conclusions}\label{sec:conclusions}
 
In this manuscript we analyze recent progresses on the understanding of the dynamics in inference problems using the tools developed in statistical physics. The attention is focused on the spiked matrix-tensor model (planted spherical mixed $p$-spin model) as a prototypical example of inference. The results on this model unveil unexpected behaviours of the dynamics and explain them from both a dynamical and a geometrical perspective. The techniques briefly summarized in the work can be extended to other models and some of the findings can be verified numerically. These are thrilling directions that we hope to pursue in the future.

\section*{Acknowledgments}

We thank G.~Biroli, C.~Cammarota, F.~Krzakala and P.~Urbani for the collaborations that led to these results and G. Bassignana, S.~Goldt and O.~Scarlatella for reading the draft of the manuscript. We acknowledge funding from the ERC under the European Union’s Horizon 2020 Research and Innovation Programme Grant Agreement 714608-SMiLe.

\end{document}